\documentclass[letterpaper, 10 pt, conference]{ieeeconf}
\IEEEoverridecommandlockouts
\usepackage[vlined, ruled, boxed, linesnumbered]{algorithm2e}

\SetCommentSty{mycommfont}

\SetKwProg{Function}{}{}{}

\usepackage{graphicx}
\usepackage{graphics}
\usepackage{times}
\usepackage{amsmath}
\usepackage{amssymb}
\usepackage{float}
\usepackage{url}
\usepackage{subcaption}
\usepackage{caption}
\usepackage{multirow}
\usepackage{gensymb}
\usepackage{epstopdf}
\usepackage{setspace} 
\usepackage[noend]{algpseudocode}
\usepackage{color}
\usepackage{cite}
\usepackage{booktabs}
\usepackage{stackengine}
\usepackage[font={small}]{caption}

\usepackage{tikz}
\usepackage{textcomp}
\usepackage{lipsum}

\title{\LARGE \bf
Robust Place Recognition using an Imaging Lidar
}

\author{Tixiao Shan, Brendan Englot, Fábio Duarte, Carlo Ratti, and Daniela Rus
\thanks{
\scriptsize{
T. Shan, F. Duarte and C. Ratti are with the Department of Urban Studies and Planning, Massachusetts Institute of Technology, USA, {\tt\scriptsize \{shant, fduarte, ratti\}@mit.edu}. \newline
\indent\; B. Englot is with the Department of Mechanical Engineering, Stevens Institute of Technology, USA, {\tt\scriptsize benglot@stevens.edu}. \newline
\indent\; T. Shan and D. Rus are with the Computer Science \& Artificial Intelligence Laboratory, Massachusetts Institute of Technology, USA, {\tt\scriptsize \{shant, rus\}@mit.edu}.} }
}

\begin{document}

\maketitle
\thispagestyle{empty}
\pagestyle{empty}


\begin{abstract}
We propose a methodology for robust, real-time place recognition using an imaging lidar, which yields image-quality high-resolution 3D point clouds. Utilizing the intensity readings of an imaging lidar, we project the point cloud and obtain an intensity image. ORB feature descriptors are extracted from the image and encoded into a bag-of-words vector. The vector, used to identify the point cloud, is inserted into a database that is maintained by DBoW for fast place recognition queries. The returned candidate is further validated by matching visual feature descriptors. To reject matching outliers, we apply PnP, which minimizes the reprojection error of visual features' positions in Euclidean space with their correspondences in 2D image space, using RANSAC. Combining the advantages from both camera and lidar-based place recognition approaches, our method is truly rotation-invariant, and can tackle reverse revisiting and upside down revisiting. The proposed method is evaluated on datasets gathered from a variety of platforms over different scales and environments. Our implementation and datasets are available at \url{https://git.io/image-lidar}.
\end{abstract}

\section{Introduction}

Place recognition plays an important role in many mobile robotics applications, such as solving the kidnapped robot problem, localizing a robot in a known map, and maintaining the accuracy of simultaneous localization and mapping (SLAM). During the last two decades, a variety of place recognition methods have achieved great success in tackling such problems using camera, lidar, and other perceptual sensors. Camera-based place recognition methods often extract visual features from textured scenes and find candidates using a bag-of-words approach. However, such methods are subject to illumination and viewpoint change. On the other hand, lidar-based place recognition methods, which often extract local or global descriptors from a point cloud, are invariant to such changes. The long detection range and wide aperture of lidar permit the capture of many structural details of an environment. Yet such details are often discarded during descriptor extraction, which may result in false positive detections when surrounded by repeating structures. Due to the prevalence of low lidar resolution, camera-based methods cannot typically be applied to lidar data. Conversely, lidar-based methods cannot typically be applied to camera data due to a lack of structural information.

\begin{figure}[t!]
	\centering
    \begin{subfigure}{.999\columnwidth}
        \centering
        \includegraphics[width=.68\textwidth]{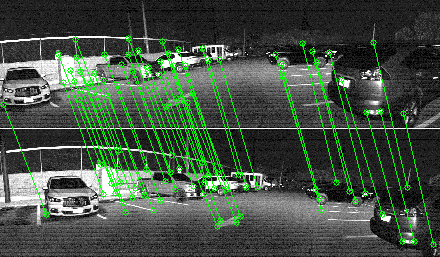}
        \includegraphics[width=.243\textwidth]{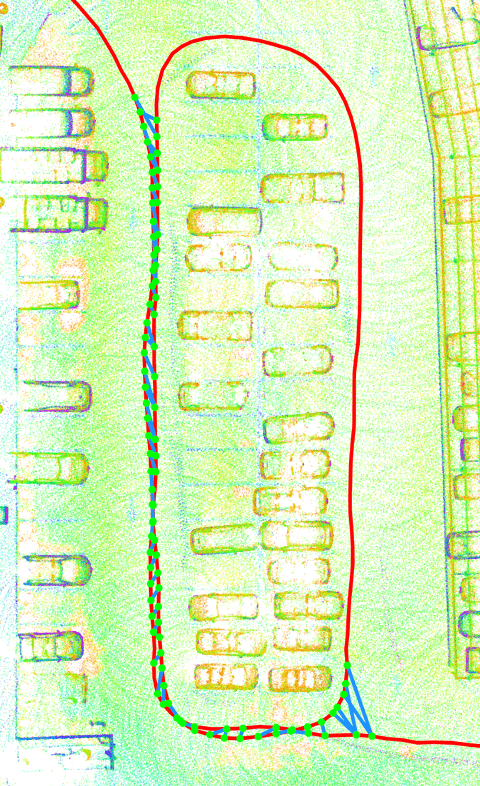}
    \end{subfigure}
	\caption{A demonstration of the proposed method applied to a mapping task. Left: a loop is found when the place is revisited. Grayscale images are intensity images projected from point clouds. Green lines connect the matched features. Right: top-view point cloud map of a parking lot. The red line indicates the traversed trajectory. Blue segments along with green dots indicate detected loop closures using our method. Note that features are extracted from the traffic arrow on the ground for place recognition.}
	\label{fig::demo-intro}
\vspace{-5mm}
\end{figure}

However, with the recent availability of high-resolution lidars, such as the Ouster OS1-128 and Velodyne VLS-128, we can begin to bridge the gap between camera-based and lidar-based place recognition methods. We refer to such high-resolution lidar that gives image-quality 3D scans as \textit{imaging lidar}. Driven by the prospects of this technology, we present a method for robust place recognition using an imaging lidar. We first project the high-resolution point cloud with intensity information onto an intensity image. We then extract Oriented FAST and rotated BRIEF (ORB) feature descriptors from the intensity image. The extracted descriptors are converted into a bag-of-words (BoW) vector, which forms a compact representation for the original point cloud. A DBoW database is built with these vectors and queried for place recognition. If a candidate is found, we match the ORB descriptors to ensure enough features can be matched between these two places. To reject matching outliers, we formulate the matching problem as an optimization problem by applying Perspective-n-Point (PnP) Random Sample Consensus (RANSAC). A representative example of our method is shown in Figure \ref{fig::demo-intro}. The main contributions of our work, which combines techniques from both camera and lidar-based place recognition methods, are as follows:
\begin{itemize}
	\item Real-time robust place recognition that is designed for imaging lidar, and to our knowledge, the first that uses projected lidar intensity images for place recognition.
	\item The proposed method, which is invariant to sensor attitude changes, is demonstrated to detect reverse revisiting, and even upside down revisiting.
	\item Our method is extensively validated with data gathered across different scales, platforms, and environments.
\end{itemize}

\begin{figure*}[t]
	\centering
    \begin{subfigure}{.205\textwidth}
        \centering
        \includegraphics[width=.85\textwidth]{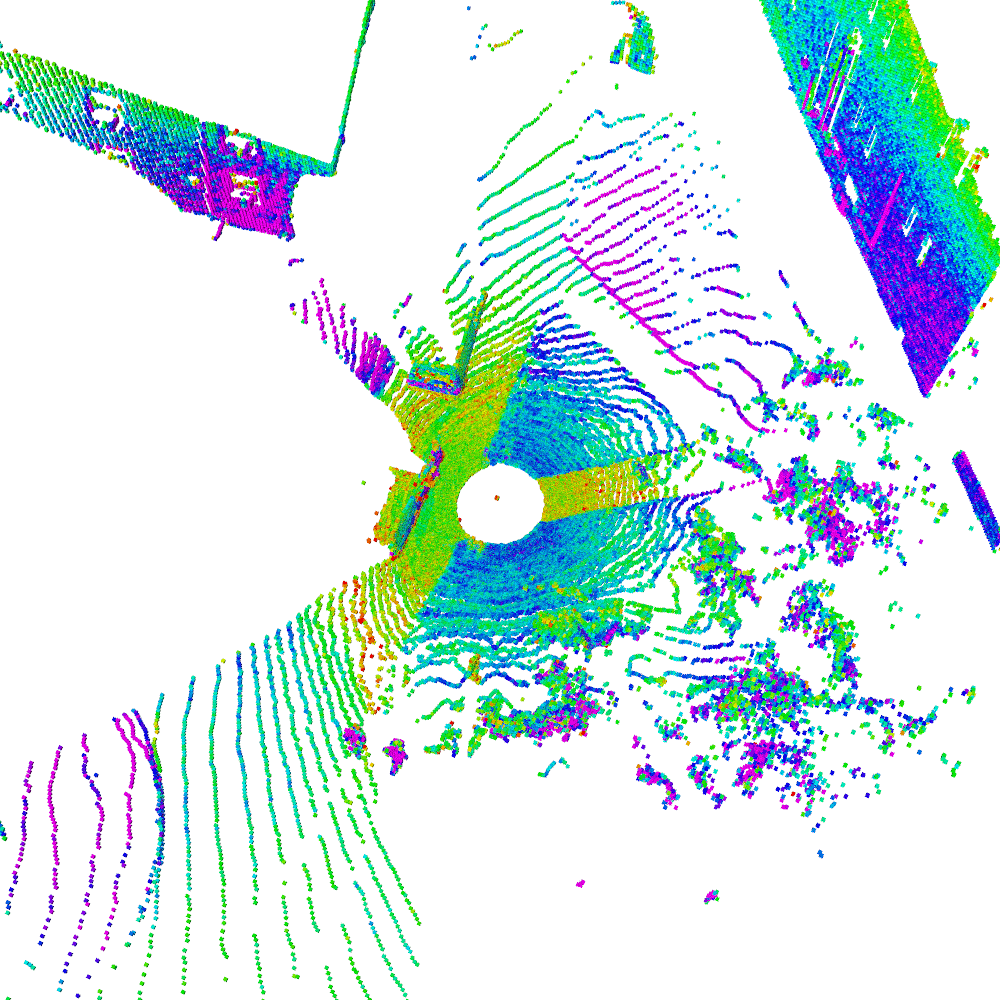}
        \caption{3D Point cloud}
    \end{subfigure}
    \begin{subfigure}{.39\textwidth}
        \centering
        \vspace{4mm}
        \includegraphics[width=.85\textwidth]{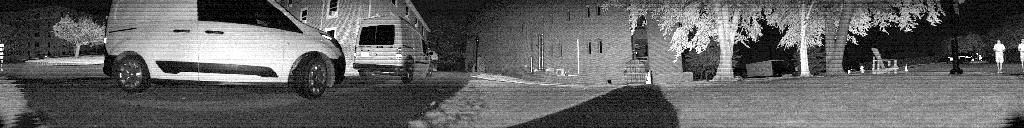}
        \\[-2mm]
        \caption{Intensity image}
        \vspace{3.5mm}
        \includegraphics[width=.85\textwidth]{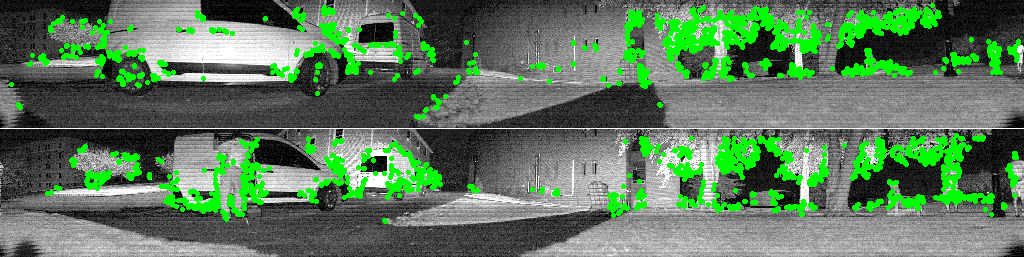}
        \\[-2mm]
        \caption{ORB features and DBoW query}
    \end{subfigure}
    \begin{subfigure}{.39\textwidth}
        \centering
        
        \includegraphics[width=.85\textwidth]{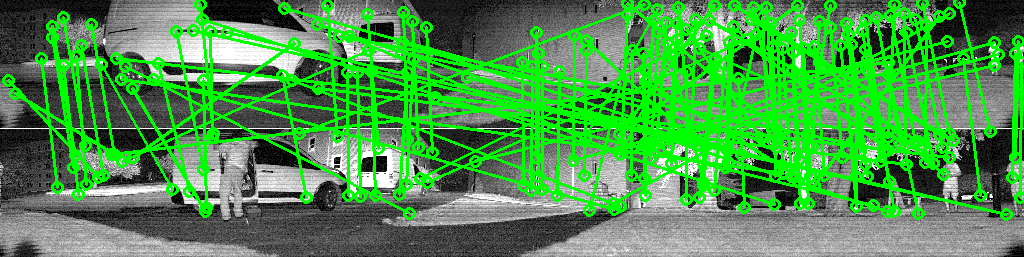}
        \\[-2mm]
        \caption{Feature matching}
        \includegraphics[width=.85\textwidth]{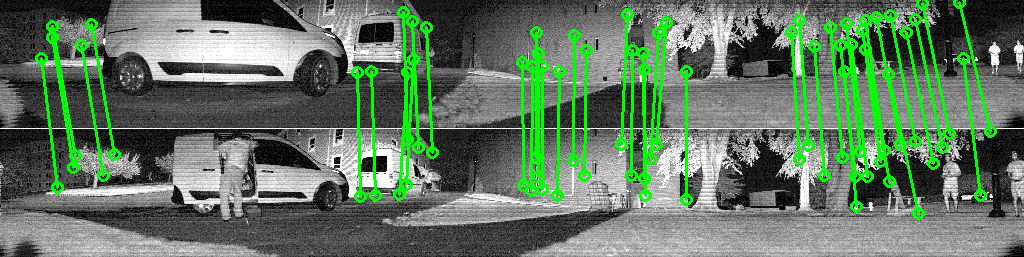}
        \\[-2mm]
        \caption{PnP RANSAC}
    \end{subfigure}
	\caption{Demonstration of the proposed methodology: (a) a high-resolution point cloud - color variation indicates intensity change; (b) the intensity image projected from the point cloud; (c) extracted ORB features (green dots) and a pair of candidates returned from a DBoW query; (d) matched ORB descriptors between two candidates; (e) matched ORB descriptors after PnP RANSAC outlier rejection.}
	\label{fig::demo-process}
	\vspace{-5mm}
\end{figure*}

\section{Related Work}

Our work draws upon concepts used in both camera-based and lidar-based place recognition methods. Due to their low hardware cost requirement and robustness in texture-rich environments, camera-based approaches have been widely used in various SLAM frameworks \cite{dvo-slam, lsd-slam, orb-slam, vins-mono} for loop closure detection. Such approaches often extract visual feature descriptors from an image and convert them into bag-of-words vectors using DBoW \cite{dbow} on a pre-trained visual vocabulary. Assuming that there exists visual overlap between images, DBoW queries the database and returns loop closure candidates based on a similarity score between vectors. Because the loop closure candidates from DBoW are prone to false detection, an extra validation step can also be introduced to reject such detection. For example, \cite{vins-mono} introduces a two-step geometric validation method, which triangulates visual features, to verify the candidates. The detection performance of camera-based methods, however, heavily depends on the environmental appearance. They are unable to offer reliable detection if the illumination and viewpoint change drastically when a place is revisited.

Lidar-based place recognition methods can be grouped into direct methods and descriptor-based methods. Though the direct methods can operate on the raw point cloud from a lidar without any pre-processing, e.g., \cite{icp, icp-variants}, their performance is sensitive to point cloud size, initial alignment, and occlusion. Therefore, we focus our discussion on descriptor-based methods, which can offer improved matching robustness. Descriptor-based methods can be categorized into local and global descriptor methods. Local descriptor methods, such as \cite{belongie2002shape}, \cite{rusu2008aligning}, and \cite{salti2014shot}, extract descriptive features from specific regions of a scan, which are then encoded into a histogram for compact representation and query. Due to the application of feature extraction in 3D space, the density of the point cloud impacts the performance of these methods. On the other hand, the recently proposed scan context (SC) \cite{sc}, intensity scan context (ISC) \cite{isc}, and lidar iris (IRIS) \cite{iris}, which are global descriptor-based methods, show superior speed and accuracy over local descriptor methods. Such methods discretize a full 3D point cloud into sectors using polar coordinates. Height or intensity information from each sector is then extracted and encoded into a 2D matrix. Thus these methods are less sensitive to the density of a point cloud. However, these methods may fail if the sensor revisits the same place with a different roll or pitch angle, which changes the signature of each sector greatly.

Similar to ISC, there is a collection of related methods that use lidar intensity information for localization and place recognition. \cite{guo2019local} and \cite{cop2018delight} rely on a custom-built scanning platform, which requires the robot to stop and scan the environment. Such a design scheme may limit their applications in real-world navigation scenarios. Deep learning-based methods, such as \cite{chen2020overlapnet} and \cite{barsan2018learning}, have incorporated lidar intensity information as input, however, we focus our discussion on methods that can be applied to a wide range of computing platforms, especially low-power embedded systems. Finally, similar to our approach, extracting visual features from lidar intensity images was proposed in \cite{barfoot2016into}, which achieves visual odometry by tracking features on a frame-to-frame basis.

In this paper, we propose a novel robust place recognition method that combines the benefits of both camera and lidar-based methods. We extract visual feature descriptors from an intensity image that is projected from the lidar point cloud. With these descriptors, we represent the point cloud using a bag-of-words vector, which is similar to lidar-based global descriptor methods. Then, we perform efficient query of these vectors using DBoW, which resembles the process of camera-based methods. At last, we verify a loop closure candidate by performing feature matching. Matched outliers are rejected by minimizing the projection error of a feature's position in Euclidean space with their correspondences in 2D image space. The method is described in detail below.

\section{Methodology}
\label{sec::system}

In this section, we describe the proposed place recognition method, intended for use with an imaging lidar. We perform a series of processing steps that includes: intensity image projection, feature extraction, DBoW query, feature matching, and PnP RANSAC. An illustrative example of each process step is shown in Figure \ref{fig::demo-process}.

\subsection{Intensity Image}
\label{sec::intensity-image}

The intensity information from a lidar represents the energy level of a return, which is generally influenced by the object surface reflectance and is invariant to ambient light. When a 3D point cloud $\mathbb{P}$ is received, we project it onto a cylindrical intensity image $\mathbb{I}$. Each valid pixel in $\mathbb{I}$ can be associated with a point in $\mathbb{P}$. The value of the pixel is determined by the intensity value of the received point. We then normalize all the pixel values to lie between 0 and 255, which essentially treats the intensity image as a grayscale image and enables us to process it with various existing image processing approaches. Pixels with no valid points associated are assigned to be zero-valued. An illustrative example of a 3D point cloud is shown in Figure \ref{fig::demo-process}(a), where color variation indicates intensity change. The resulting intensity image is shown in Figure \ref{fig::demo-process}(b), where the light and dark pixels correspond to high and low intensity values respectively.

\subsection{Feature Extraction}
\label{sec::feature-extraction}

We next perform feature extraction on the intensity image $\mathbb{I}$. Rather than assuming a fixed sensor mounting solution \cite{sc, isc, iris}, we assume the lidar sensor may undergo aggressive orientation change, which greatly extends the application scenarios of our approach. Therefore, we choose to extract ORB features \cite{orb} due to their efficiency and invariance to rotation change. ORB feature descriptors are obtained by first extracting FAST corner features \cite{fast} and then describing them using BRIEF descriptors \cite{brief}. 
Due to sensor motion, the scale of an object observed in the intensity image is a function of the distance between the sensor and the object. Similarly, the apparent orientation of the object is also subject to sensor orientation. To increase extraction robustness at various scales and orientations, we apply an eight-level image pyramid with a down-sample ratio of 1.2 to obtain eight intensity images at different resolutions. ORB features are detected using the FAST algorithm in each of the images. The orientation of a feature is determined by computing the intensity change in a circular region that is centered at the feature. The BRIEF algorithm is then used to convert a corner feature to a descriptor. We extract a total number of $N_{bow}$ ORB feature descriptors, which are denoted $\mathbb{O}$. Note that since we associate each 3D point in $\mathbb{P}$ to each pixel in $\mathbb{I}$, every feature descriptor in $\mathbb{O}$ is also associated with a 3D point in $\mathbb{P}$. An example of the extracted ORB features overlaid on an intensity image is shown in Fig. \ref{fig::demo-process}(c).

\subsection{DBoW Query}
\label{sec::dbow-euqry}

We utilize DBoW \cite{dbow} to convert the ORB feature descriptors $\mathbb{O}$ into a bag-of-words vector using the visual vocabulary proposed in \cite{orb-slam}. Thus, the 3D point cloud is now efficiently represented using a sparse bag-of-words vector, which is used to build a database with DBoW. When a new bag-of-words vector is received, we query the database by measuring the similarity between the new vector and the previous vectors using the $L1$ distance. If the similarity between two vectors is larger than a threshold $\lambda_{bow}$, we assume a potential revisit candidate is found. The new bag-of-words vector is inserted into the database after the query. Denoting the timestamps at the current and previous vectors as $i$ and $j$, we send $\mathbb{O}_i$ and $\mathbb{O}_j$ to the processes described in the following sections for further verification. A matched candidate returned by DBoW is shown in Figure \ref{fig::demo-process}(c), where the top and bottom images represent $\mathbb{I}_i$ and $\mathbb{I}_j$ respectively.

\subsection{Feature Matching}
\label{sec::feature-matching}

Usually, the candidates from a DBoW query consist of many false detections. To validate a detection, we match the descriptors from $\mathbb{O}_i$ and $\mathbb{O}_j$. We note that matching every descriptor in $\mathbb{O}_i$ and $\mathbb{O}_j$ is not only computationally expensive, but also often results in numerous false matches. To increase the matching success rate, we rank all the descriptors of $\mathbb{O}_i$ in descending order based on their corner scores \cite{harris}. The first $N_s$ descriptors, where $N_s < N_{bow}$, with the largest corner scores are selected and denoted as $\mathbf{O}_i: \mathbf{O}_i \subset \mathbb{O}_i$. For each descriptor in $\mathbf{O}_i$, we find its best match in $\mathbb{O}_j$. The distance between two descriptors is calculated using the Hamming distance. The matched descriptors are then ranked in ascending order based on their Hamming distance. At last, we introduce a distance test to reject false matches - only matches with Hamming distance less than $\lambda_h$ are retained for further validation. We set $\lambda_h$ to twice the smallest Hamming distance of all current matches. In practice, we find that our distance test performs better than \cite{lowe2004distinctive}, which rejects many true positive matches. The matched descriptors are denoted as $O_i : O_i \subset \mathbf{O}_i \subset \mathbb{O}_i$ and $O_j : O_j \subset \mathbb{O}_j$ respectively, where we have $\lVert O_i \rVert = \lVert O_j \rVert$. 

An example of matched descriptors is shown in Figure \ref{fig::demo-process}(d). Note that there are still many false positive matches across the images. Though we can choose a smaller $\lambda_h$ to filter such matches, many true positive matches may be rejected as well. If the number of successful matches after the distance test is larger than $N_m$, we then proceed to the PnP RANSAC technique described next in Section \ref{sec::pnp-ransac}.

\subsection{PnP RANSAC}
\label{sec::pnp-ransac}

If the candidate returned by a DBoW query is incorrect, we may still obtain enough matches from the process described in  Section \ref{sec::feature-matching}. To further validate the candidate, we formulate the validation problem as a PnP problem \cite{pnp}. Knowing the 3D Euclidean position of features in $O_i$ and the 2D image position of features in $O_j$, PnP minimizes the reprojection error of the 3D points and their 2D correspondences, and estimates the relative sensor pose between $i$ and $j$. However, PnP is prone to errors due to false matches in $O_i$ and $O_j$, which is shown in Figure \ref{fig::demo-process}(d). To increase the robustness of PnP, we utilize RANSAC \cite{ransac} here to reject outliers among the matches. Figure \ref{fig::demo-process}(e) shows the correct feature matches after outlier rejection. Note that the matched features surrounding the observed person, who is not observed near the van in the latest frame, are rejected after performing PnP RANSAC. If the number of inliers exceeds $N_p$, we treat this candidate as a correct detection. The estimation of relative sensor pose between $i$ and $j$, which is a byproduct of PnP, can also be used in a full SLAM framework to facilitate frame-to-frame registration.

\begin{figure*}[ht]
	\centering
    \begin{subfigure}{.99\textwidth}
        \centering
        \includegraphics[width=.999\textwidth]{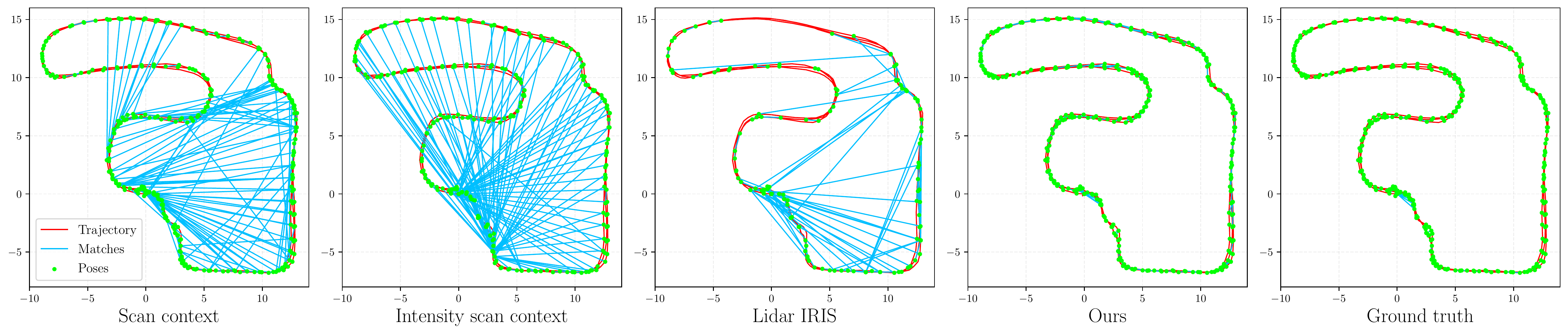}
        \\[-2mm]
        \caption{Indoor dataset}
    \end{subfigure}
    \begin{subfigure}{.999\textwidth}
        \centering
        \includegraphics[width=.999\textwidth]{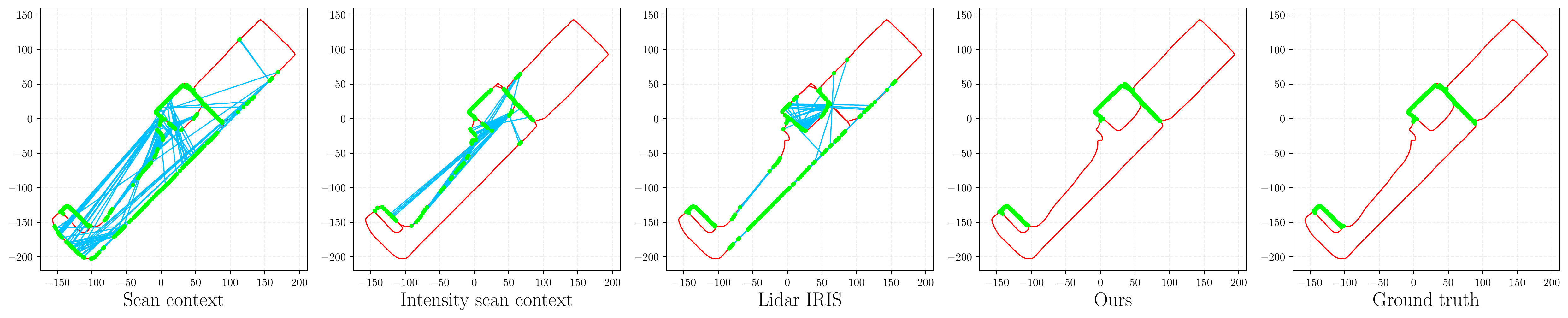}
        \\[-2mm]
        \caption{Handheld dataset}
    \end{subfigure}
    \begin{subfigure}{.999\textwidth}
        \centering
        \includegraphics[width=.999\textwidth]{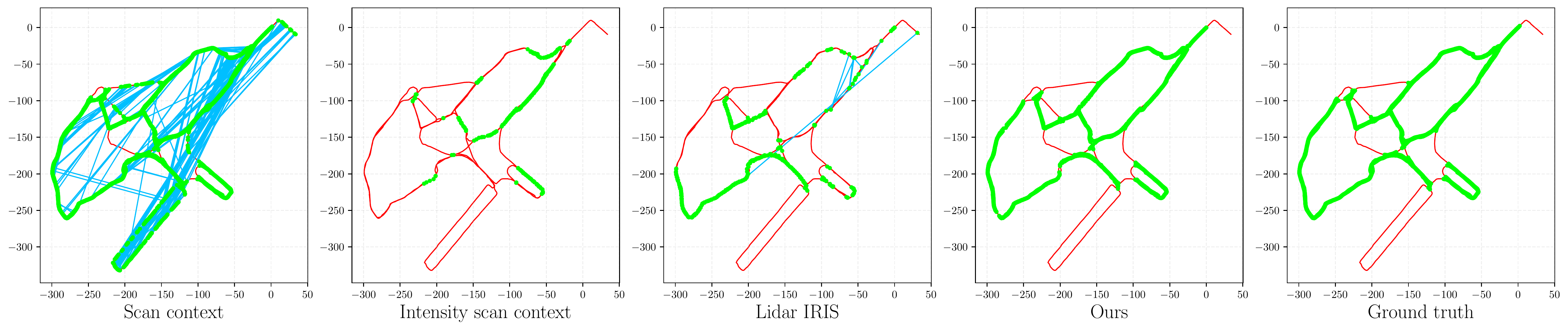}
        \\[-2mm]
        \caption{Jackal dataset}
    \end{subfigure}
	\caption{Detected loop closures of each method. The trajectory of the dataset is colored red. The green dots and blue segments indicate the reported loop closure matches, where the green dots represent the positions of loop closures. Axis units are in meters.}
	\label{fig::benchmarking-map}
	\vspace{-5mm}
\end{figure*}

\section{Experiments}

We now describe a series of experiments to quantitatively analyze the proposed method. The sensor used in this paper is the Ouster OS1-128 imaging lidar. The horizontal and vertical field-of-view of the sensor are 360$^\circ$ and 45$^\circ$ respectively. The resolution of the sensor in both directions is 0.35$^\circ$ when it operates at 10Hz, thus resulting in an intensity image with a resolution of 1024 by 128. We compare the proposed method with SC \cite{sc}, ISC \cite{isc}, and IRIS \cite{iris}. All the methods are implemented in C++ and executed on a laptop equipped with an Intel i7-10710U 1.1GHz CPU.

For validation, we gathered three different datasets across various scales, platforms and environments. These datasets are referred to as \textit{Indoor}, \textit{Handheld}, and \textit{Jackal}, respectively. The lidar scans from the OS1-128 are registered using LIO-SAM \cite{lio-sam}, which is a tightly-coupled lidar-inertial odometry framework built atop a factor graph. Similar to the previous implementation of SC using \cite{lego-loam}, we associate each place detection to a node in the factor graph of LIO-SAM, which allows each method's performance to be validated. In addition to evaluating each place recognition method atop the LIO-SAM solution, we also compare against a ground truth solution in which all possible loop closures more than 30 seconds apart are identified. In our comparison, we use the default parameters from the available implementations of SC, ISC, and IRIS. The parameters of our method are chosen as: $\lambda_{bow}=0.015$, $N_s=500$, $N_{bow}=2500$, $N_m=15$, $N_p=15$ for all experiments. Supplementary details of the experiments performed can be found at the link below\footnote{\url{https://youtu.be/yH1hLBFaNoI}}.

\vspace{-1.5mm}

\begin{figure*}[ht]
	\centering
    \begin{subfigure}{.9\textwidth}
        \centering
        \includegraphics[width=.42\textwidth]{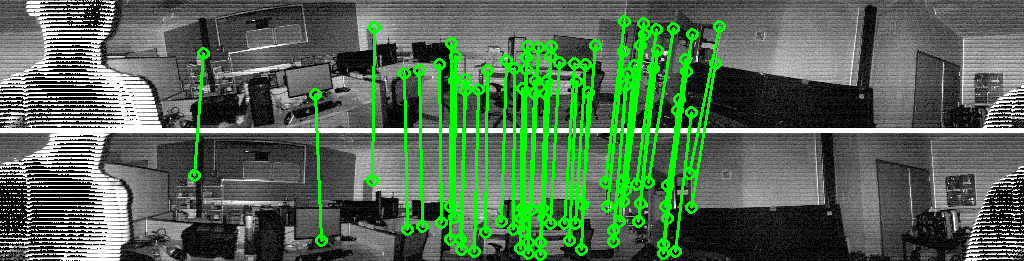}
        \includegraphics[width=.42\textwidth]{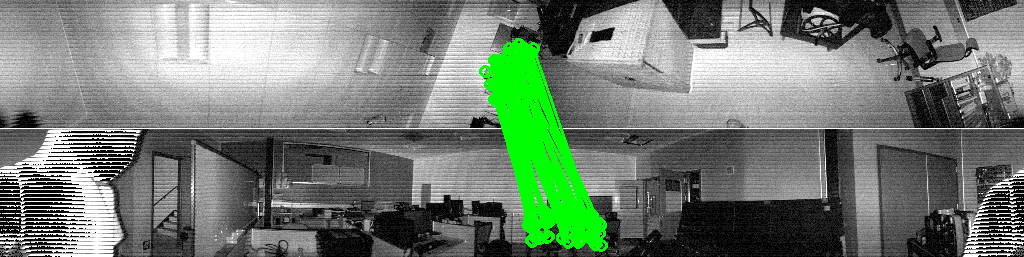}
        \\[0.5mm]
        \includegraphics[width=.42\textwidth]{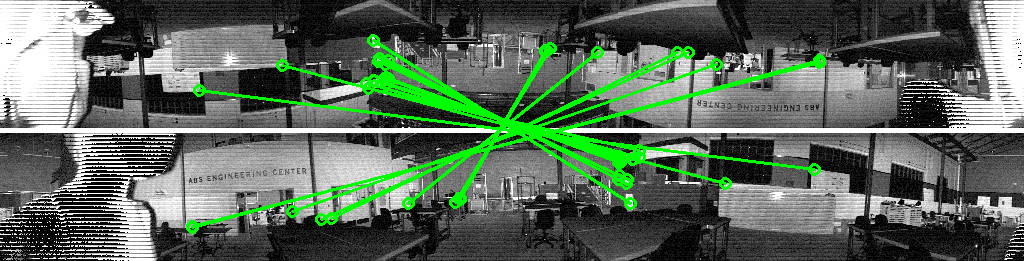}
        \includegraphics[width=.42\textwidth]{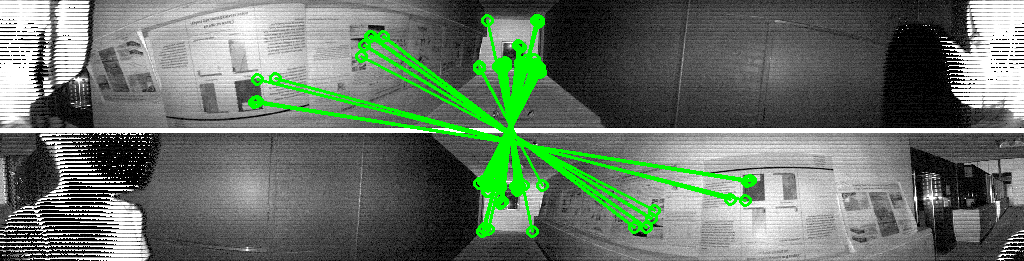}
        \\[-1.5mm]
        \caption{Indoor dataset}
    \end{subfigure}
    \\[1mm]
    
    \begin{subfigure}{.9\textwidth}
        \centering
        \includegraphics[width=.42\textwidth]{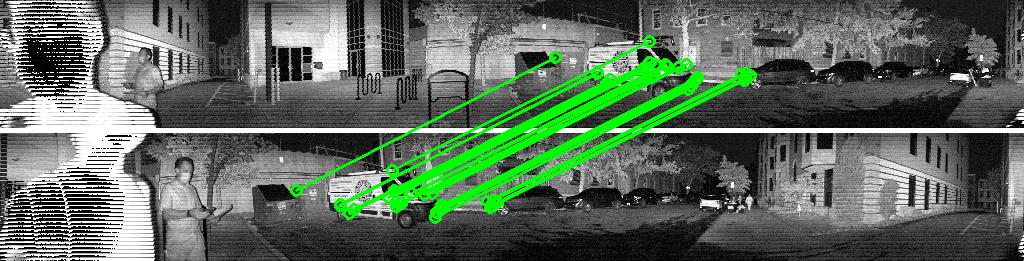}
        \includegraphics[width=.42\textwidth]{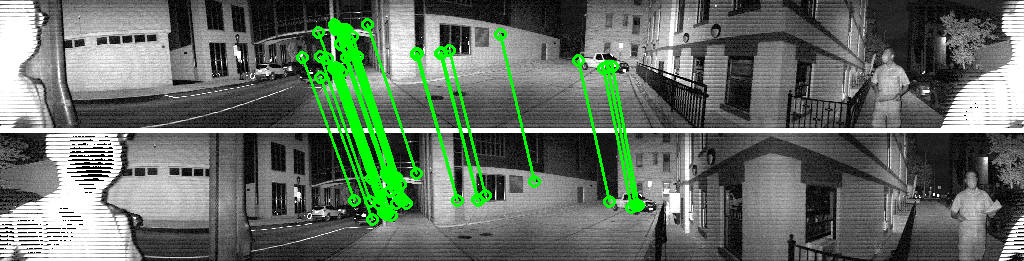}
        \\[0.5mm]
        \includegraphics[width=.42\textwidth]{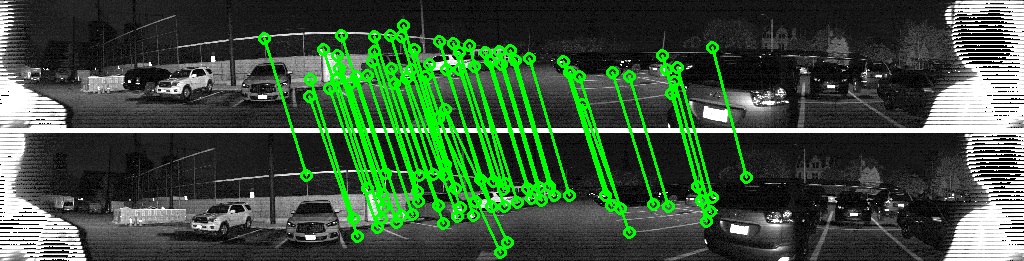}
        \includegraphics[width=.42\textwidth]{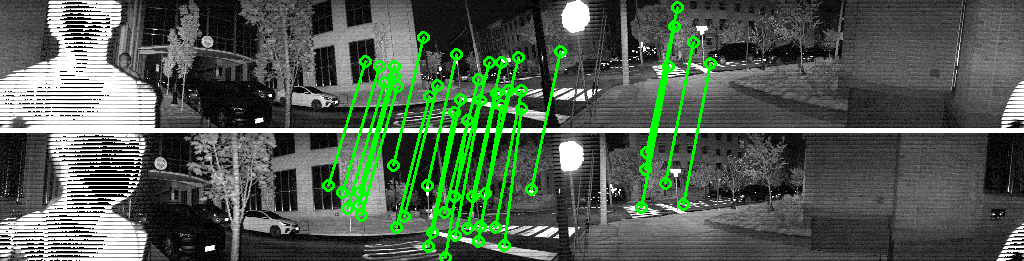}
        \\[-1.5mm]
        \caption{Handheld dataset}
    \end{subfigure}
    \\[1mm]
    
    \begin{subfigure}{.9\textwidth}
        \centering
        \includegraphics[width=.42\textwidth]{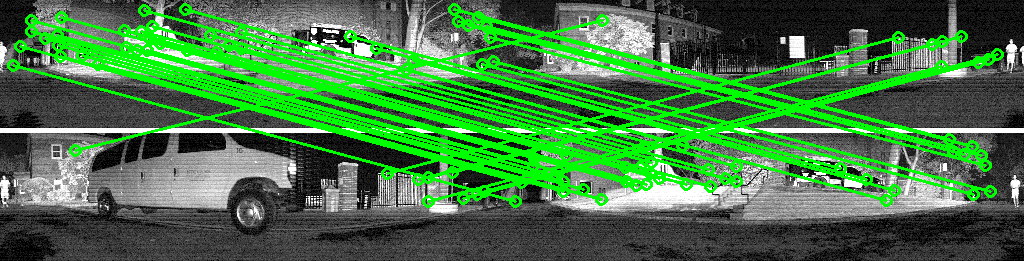}
        \includegraphics[width=.42\textwidth]{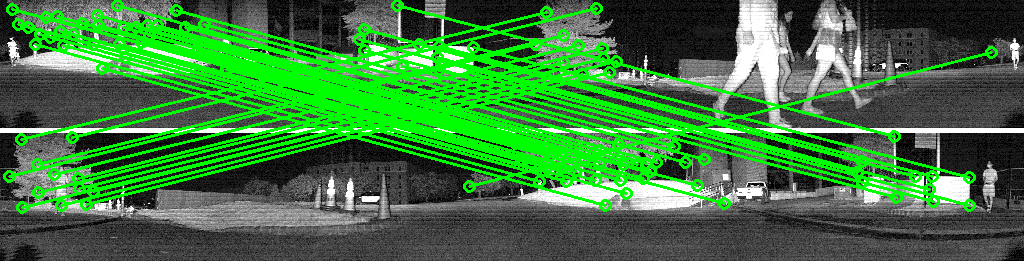}
        \\[0.5mm]
        \includegraphics[width=.42\textwidth]{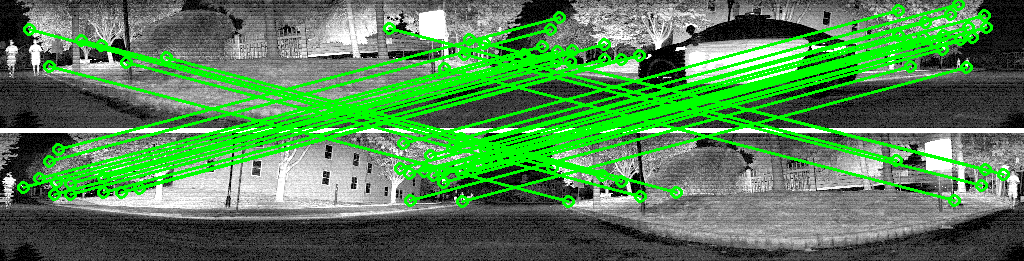}
        \includegraphics[width=.42\textwidth]{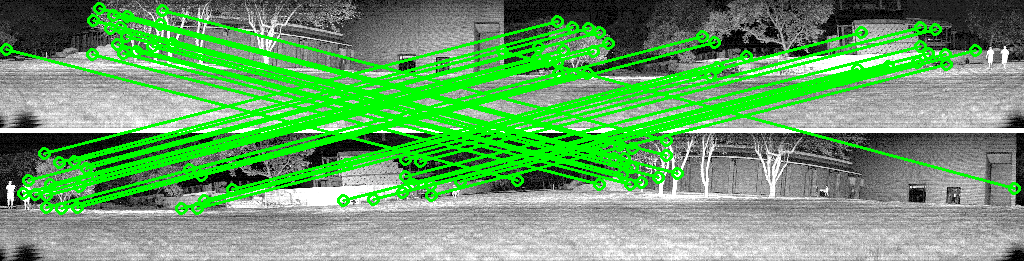}
        \\[-1.5mm]
        \caption{Jackal dataset}
    \end{subfigure}
    
	\caption{Twelve representative loop closure detection examples using our method. For each example, the top and bottom row images indicate intensity images captured at the current and previous times, respectively. Matched ORB features are connected using green lines.}
	\label{fig::dataset-demo}
	\vspace{-5mm}
\end{figure*}

\subsection{Indoor Dataset}

The \textit{Indoor} dataset is gathered by an operator carrying the sensor walking in an indoor environment, which passes through doors, corridors, and areas populated with furniture. During data-gathering, the operator follows the same trajectory three times, which start and finish at the same location. When traversing the environment for the third time, the operator turns the sensor completely upside down. Ideally, a robust loop closure method should start reporting detections when the environment is passed the second and third time. 

The detected loop closures of each method are shown in Figure \ref{fig::benchmarking-map}(a). The LIO-SAM trajectory is colored red. The green dots and blue segments indicate the loop closure matches, where the green dots represent the position of the node in the factor graph of LIO-SAM. If the position between the matched nodes is less than 2m, we consider this detection a true positive, otherwise a false positive. The \textit{Indoor} dataset has 245 ground truth loop closures. Due to the detection mechanisms of SC, ISC, and IRIS, fine details of the environment are discarded in the process of obtaining their point cloud descriptors. Thus, many false positives are reported by these methods, especially when the trajectory is traversed for the third time with the sensor upside down. 

Four representative loop closures detected by our method are shown in Figure \ref{fig::dataset-demo}(a). For each detection, the top and bottom row images indicate intensity images captured at the current and previous times. Matched ORB features are connected using green lines. The second example shows a detection when we rotate the sensor close to 90 degrees around its forward axis. The third and fourth examples show our method detecting loop closures with the sensor upside down, which is a 180 degree rotation about its forward axis. Note that in the fourth example, our method extracts features from posters hanging in the corridor to aid detection.

\begin{figure}[ht]
	\centering
    \includegraphics[width=.99\columnwidth]{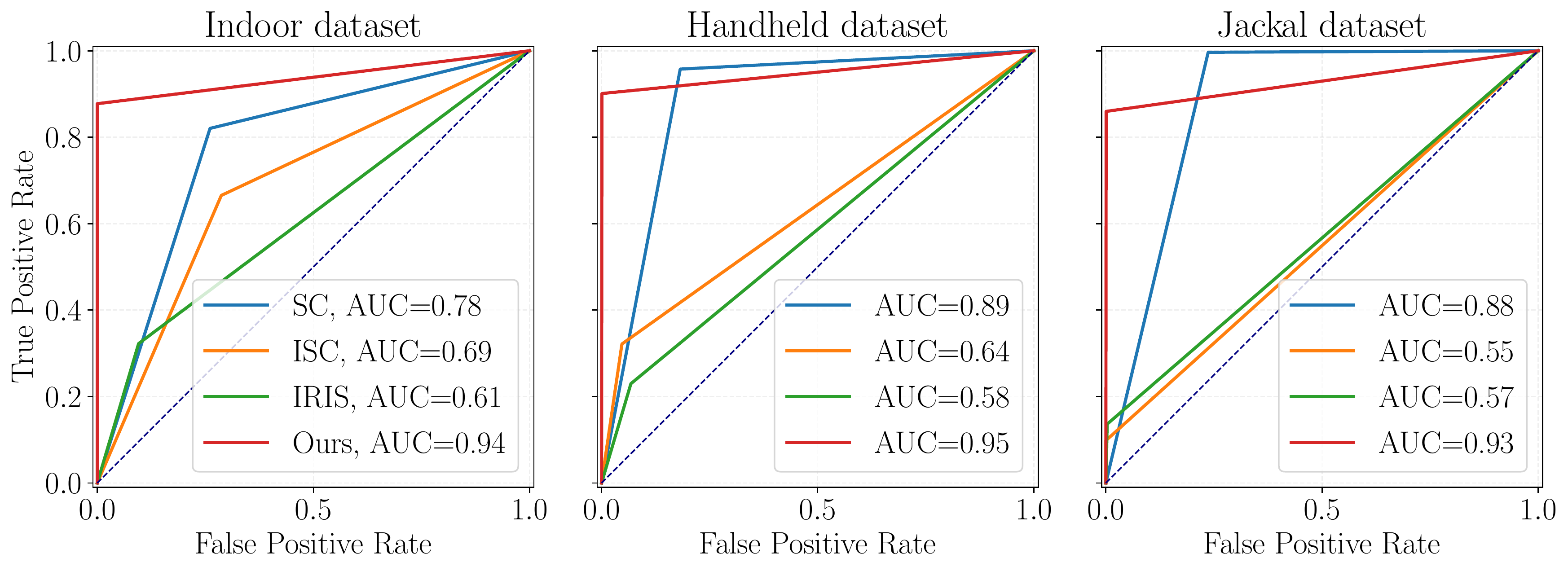}
	\caption{ROC curves and AUC for all competing methods. The results are obtained by comparing the reported loop closure detection with the ground truth loop closures. Among all the methods, the proposed method achieves the highest AUC over various datasets.}
	\label{fig::roc-curve}
	\vspace{-3mm}
\end{figure}

The number of true and false positives reported by each method is shown in Table \ref{tab::benchmarking-all}. Among all the detections recorded, the positive detection rates of SC, ISC, IRIS, and our method are 58\%, 51\%, 47\%, and 100\% respectively. We also use receiver operating characteristic (ROC) curves to benchmark the detection accuracy of each method. The ROC curves, which are shown in Figure \ref{fig::roc-curve}(a), plot the true positive rate against the false positive rate. The area under the curve (AUC) is provided for comparison of prediction accuracy.

\subsection{Handheld Dataset}

The $Handheld$ dataset is gathered in an outdoor environment with the operator walking with the sensor. This dataset features urban structures and vegetation, moving cars and pedestrians. Since it is not mounted on a fixed platform, the sensor undergoes aggressive attitude change. The distance between matches for considering true positive detection is increased to 4m due to environment scale change.

The reported loop closure detections of each method are shown in Fig. \ref{fig::benchmarking-map}(b). Note that SC, ISC, and IRIS report many false positives when we pass through two long streets, which consist of many repetitive scenes (shown in the lower left of the figure). Our method, on the other hand, can reject false detections by utilizing the fine details of the environment. Four examples of detected loop closures by our method are shown in Fig. \ref{fig::dataset-demo}(b), where cars, windows, and street markings are utilized for detection. As shown in Table \ref{tab::benchmarking-all}, our method achieves a 98\% true positive detection rate, as opposed to 54\% for SC, 60\% for ISC, and 31\% for IRIS. We also achieve the highest AUC of all methods in Fig. \ref{fig::roc-curve}(b).


\subsection{Jackal Dataset}

In the \textit{Jackal} dataset, we mount the sensor on a Clearpath Jackal unmanned ground vehicle (UGV), driving the UGV on asphalt roads, concrete and brick sidewalks, and ground covered by grass and soil. We mainly test the reverse loop closure detection capability  of all methods. Though SC is able to detect 1429 out of 1447 ground truth loop closures, it reports 409 false positives, which accounts for 22\% of all detections. ISC and IRIS report significantly fewer detections compared with SC and our method. Our method detects 1245 loop closures with 98\% of them being true positives. Again, our method achieves the highest AUC in Figure \ref{fig::roc-curve}(c).

Representative reverse loop closure detection examples (involving traversal in opposite directions) are shown in Fig. \ref{fig::dataset-demo}(c). Our method extracts features primarily from trees and buildings to support place recognition. Many moving vehicles and pedestrians are observed during the data-gathering process, and our proposed method rejects the matched features from dynamic objects using the technique discussed in Sec. \ref{sec::pnp-ransac}. In the fourth example of Fig. \ref{fig::dataset-demo}(c), the ground surrounding the sensor is completely covered by grass.

\subsection{Runtime Benchmarking}

The computation time per query averaged over each dataset, for each method, is shown in the last column of Table \ref{tab::benchmarking-all}. SC is the most time-efficient method due to its introduction of a ring key search algorithm for fast database query. Though the point cloud descriptor comparison times for ISC and IRIS are similar to SC, their computation times increase dramatically when they are applied to a full SLAM framework. ISC naively compares the current descriptor with all the descriptors in the database during a query, thus its computation time grows unbounded. Though IRIS implements a similar search algorithm as SC for query, its efficiency is not ideal due to the design of the search key, which has a dimension of 80 as opposed to 20 in SC. Though our method runs slower than SC, it is significantly faster than ISC and IRIS, while achieving the highest true positive detection accuracy. It's worth noting that average DBoW query time for the three datasets is 22.2 ms, 33.1 ms, and 58.8 ms respectively, which increases as the size of the database increases. The computation time for the remaining components of our method is similar across all datasets.

\begin{table}[ht]
\caption{Quantitative results of competing methods}
\label{tab::benchmarking-all}
\centering
\resizebox{0.99\columnwidth}{!}
{
    \begin{tabular}{@{}cccccc@{}}
    \toprule
    Dataset & Method & \begin{tabular}[c]{@{}c@{}}Detected loops\end{tabular} & \begin{tabular}[c]{@{}c@{}}True positives\end{tabular} & \begin{tabular}[c]{@{}c@{}}False positives\end{tabular} & \begin{tabular}[c]{@{}c@{}}Time (ms)\end{tabular} \\ 
    \midrule
    \multirow{4}{*}{\begin{tabular}[c]{@{}c@{}}Indoor\\ (245 loops)\end{tabular}} & SC & 231 & 134 (58\%) & 97 (42\%) & 6.17 \\
     & ISC & 196 & 100 (51\%) & 96 (49\%) & 164.8 \\
     & IRIS & 90 & 42 (47\%) & 48 (53\%) & 253.4 \\
     & Ours & 215 & 215 (100\%) & 0 (0\%) & 49.5 \\
    \midrule
    \multirow{4}{*}{\begin{tabular}[c]{@{}c@{}}Handheld\\ (283 loops)\end{tabular}} & SC & 499 & 271 (54\%) & 228 (46\%) & 6.49 \\
     & ISC & 150 & 90 (60\%) & 60 (40\%) & 417.0 \\
     & IRIS & 150 & 47 (31\%) & 103 (69\%) & 382.4 \\
     & Ours & 256 & 252 (98\%) & 4 (2\%) & 65.7 \\
    \midrule
    \multirow{4}{*}{\begin{tabular}[c]{@{}c@{}}Jackal\\ (1447 loops)\end{tabular}} & SC & 1838 & 1429 (78\%) & 409 (22\%) & 9.19 \\
     & ISC & 142 & 134 (94\%) & 8 (6\%) & 630.4 \\
     & IRIS & 202 & 168 (83\%) & 34 (17\%) & 423.5 \\
     & Ours & 1245 & 1226 (98\%) & 19 (2\%) & 93.2 \\
     \bottomrule
    \end{tabular}
}
\vspace{-3mm}
\end{table}

\subsection{Lidar Resolution Benchmarking}

Finally, we provide detection results for our method using down-sampled intensity images. The image resolutions tested are 1024 by 64, 1024 by 32, and 1024 by 16, which are equivalent to using a lidar with 64, 32, and 16 channels respectively. As shown in Table \ref{tab::benchmarking-channel}, the detection rate when using fewer lidar channels decreases significantly. This is because the number of extracted ORB features from a low-resolution intensity image is limited, and the performance of DBoW query and feature matching deteriorates accordingly. Our method is clearly sensitive to the available lidar resolution, and most suitable for use with hi-res imaging lidar.

\vspace{-1mm}

\begin{table}[ht]
\caption{Detection with differing lidar resolution}
\label{tab::benchmarking-channel}
\centering
\resizebox{0.99\columnwidth}{!}
{
    \begin{tabular}{ccccc}
    \toprule
    Dataset & \begin{tabular}[c]{@{}c@{}}Lidar channels\end{tabular} & \begin{tabular}[c]{@{}c@{}}Detected loops\end{tabular} & \begin{tabular}[c]{@{}c@{}}True positives\end{tabular} & \begin{tabular}[c]{@{}c@{}}False positives\end{tabular} \\
    \midrule
    \multirow{3}{*}{\begin{tabular}[c]{@{}c@{}}Indoor\\ (245 loops)\end{tabular}} & 64 & 181 & 177 (98\%) & 4 (2\%) \\
     & 32 & 75 & 75 (100\%) & 0 (0\%) \\
     & 16 & 3 & 3 (100\%) & 0 (0\%) \\
    \midrule
    \multirow{3}{*}{\begin{tabular}[c]{@{}c@{}}Handheld\\ (283 loops)\end{tabular}} & 64 & 236 & 234 (99\%) & 2 (1\%) \\
     & 32 & 104 & 103 (99\%) & 1 (1\%) \\
     & 16 & 6 & 5 (83\%) & 1 (17\%) \\
    \midrule
    \multirow{3}{*}{\begin{tabular}[c]{@{}c@{}}Jackal\\ (1447 loops)\end{tabular}} & 64 & 1040 & 1022 (98\%) & 18 (2\%) \\
     & 32 & 435 & 428 (98\%) & 7 (2\%) \\
     & 16 & 35 & 34 (97\%) & 1 (3\%) \\
    \bottomrule
    \end{tabular}
}
\vspace{-3mm}
\end{table}

\vspace{-1mm}

\section{Conclusions and Discussion}

We propose a novel methodology for place recognition using an imaging lidar, which demonstrates robustness in a variety of settings. Our method combines advantages from both camera and lidar-based place recognition approaches. Similar to camera-based methods, we extract ORB feature descriptors from the intensity image projected from a 3D point cloud. We utilize DBoW to represent the point clouds using bag-of-words vectors and to perform place recognition queries, which is similar to lidar-based global descriptor methods. Upon receiving a candidate from a query, we conduct ORB descriptor matching to verify its legitimacy. The outliers among the matched descriptors are rejected using PnP RANSAC. The proposed method is evaluated on datasets gathered in both indoor and outdoor environments at different scales. The results show that our method achieves higher accuracy and robustness than other lidar-based place recognition methods.

Because the lidar scan points are not received simultaneously, a lidar scan usually suffers from motion distortion. We did not consider this effect in our experiments due to the slow motion of our sensor. However, the performance impact of lidar distortion on place recognition for high-speed platforms is an important area of interest for future work.

We are aware that the KITTI dataset \cite{kitti} is used in \cite{sc}, \cite{isc}, and \cite{iris} for benchmarking. However, the lidar, Velodyne HDL-64e, used in the KITTI dataset features nonlinearly distributed channels along its spinning axis. Depending on the sensor attitude, the same object may have different appearances at different vertical locations of the intensity image. We are unable to extract consistent ORB feature descriptors for DBoW query or feature matching. Therefore, we did not include results using the KITTI dataset.

\section*{Acknowledgement}

This work was supported by Amsterdam Institute for Advanced Metropolitan Solutions, Amsterdam, the Netherlands.

\bibliographystyle{IEEEtran}
\bibliography{ICRA_2021_LOOP}
\end{document}